# SDIGLM: Leveraging Large Language Models and Multi-Modal Chain of Thought for Structural Damage Identification

Yunkai Zhang[1,2], Shiyin Wei[1,2], Yong Huang[1,2*], Yawu Su[3], Shanshan Lu[3], Hui Li[1,2]

[1.] Key Lab of Smart Prevention and Mitigation of Civil Engineering Disasters of the Ministry of Industry and Information Technology, School of Civil Engineering, Harbin Institute of Technology, Harbin, 150090, China

[2.] Key Lab of Structures Dynamic Behavior and Control of the Ministry of Education, Harbin Institute of Technology, Harbin, 150090, China

[3.] CSCEC Digital Technology Co., Ltd., Beijing 100000, China

**Abstract:** Existing computer vision(CV)-based structural damage identification models demonstrate notable accuracy in categorizing and localizing damage. However, these models present several critical limitations that hinder their practical application in civil engineering(CE). Primarily, their ability to recognize damage types remains constrained, preventing comprehensive analysis of the highly varied and complex conditions encountered in real-world CE structures. Second, these models lack linguistic capabilities, rendering them unable to articulate structural damage characteristics through natural language descriptions. With the continuous advancement of artificial intelligence(AI), large multi-modal models(LMMs) have emerged as a transformative solution, enabling the unified encoding and alignment of textual and visual data. These models can autonomously generate detailed descriptive narratives of structural damage while demonstrating robust generalization across diverse scenarios and tasks. This study introduces SDIGLM, an innovative LMM for structural damage identification, developed based on the open-source VisualGLM-6B architecture. To address the challenge of adapting LMMs to the intricate and varied operating conditions in CE, this work integrates a U-Net-based semantic segmentation module to generate defect segmentation maps as visual Chain of Thought(CoT). Additionally, a multi-round dialogue fine-tuning dataset is constructed to enhance logical reasoning, complemented by a language CoT formed through prompt engineering. By leveraging this multi-modal CoT, SDIGLM surpasses general-purpose LMMs in structural damage identification, achieving an accuracy of 95.24% across various infrastructure types. Moreover, the model effectively describes damage characteristics such as hole size, crack direction, and corrosion severity.

## 1. Introduction

Civil Engineering (CE) structures such as buildings, bridges, and roads will inevitably sustain damage from the external environment during their long-term service. Such damages include cracks, holes, concrete spalling, reinforcement exposure, corrosion, and more. These damages not only impair the aesthetic appearance of the structures but may also have a significant impact on their safety and stability. In the traditional process of civil engineering operation and maintenance, inspections mainly rely on visual checks by inspectors. This approach is not only time-consuming and labor-intensive but also highly vulnerable to the subjective biases of the inspectors.

In recent years, computer vision algorithms have shown great advantages in automatic damage detection, greatly improving the detection efficiency and accuracy of structural damage recognition. These methods mainly include image classification, target detection, semantic segmentation and instance segmentation. These methods can effectively recognize the damage categories and the damage areas from both qualitative and quantitative perspectives. Structure damage classification methods[1-3] aim to distinguish different damage categories. Most classification models are based on convolutional neural network (CNN)[4] and the parameters or network structure is continuously optimized. Bhattacharya et al.[3] introduced multi-scale committee of attention and fine-grained feature induced attention modules into the CNN learning architecture in order to improve the classification accuracy of multi-target and multi-category concrete defects. Target detection algorithms[5-7] output both bounding boxes and category labels for target objects. Yu et al.[7] proposed a YOLOv4-FPM model integrating focus loss, pruning algorithms, and multi-scale datasets for real-time detection of bridge cracks based on unmanned aircraft. Semantic segmentation model[8-10] labels each pixel in an image to delineate damage and background. Chaiyasarn et al.[9] used CNN for image classification, and then used a full convolutional network to implement semantic segmentation based on the classification results. Compared with non-pixel-level methods, the semantic segmentation algorithm realizes the complete segmentation of defective pixels and background pixels. Li et al.[10] take post-processing methods to obtain a series of damage features, such as length, width, area, based on the segmentation results. Instance segmentation model[11, 12] combines the functionality of target detection and semantic segmentation algorithms to realize the labeling of different damages with different labeled pixel levels. Mask R-CNN[11] achieves mask segmentation by first generating candidate bounding boxes, and then by predicting the class of each candidate box and realizing the mask segmentation. Wei et al.[12] use Mask R-CNN to identify the holes on the surface of the concrete structure, based on the example segmentation results, quantifying the area and maximum diameter of the holes by determining the ratio relationship between the pixel area and the actual area.

Existing structural damage identification models demonstrate high accuracy in detecting common damage categories and regions, such as concrete cracks and holes. However, the structural environment in CE is inherently complex, encompassing diverse structural types and a wide range of damage patterns. The generalization ability of current models remains limited, often necessitating scenario-specific modeling, which significantly hinders practical applications. Additionally, these models struggle to bridge the modality gap between image-based and text-based data, lacking capabilities in natural language reasoning and generation. Consequently, they are unable to describe damage characteristics in natural language, and to retrieve structural operation and maintenance knowledge from identification results.

With Google's release of Transformer[13], the highly parallel framework greatly improves computational efficiency, and makes it possible to deep learning models an increasingly large number of parameters. In the field of natural language processing, researchers trained pre-trained large language models (LLMs) with billions to trillions of parameters using a large amount of training data and high-performance graphics cards. These models segment and encode natural language texts, and exhibit abilities such as instruction following and step-by-step reasoning that traditional small models do not possess[14]. Among them, ChatGPT[15] released by OpenAI has already demonstrated language abilities close to those of humans in tasks such as human-computer interaction, translation, and document generation.

Subsequently, in order to extend the powerful text processing capabilities of LLMs to multi-modal data such as images and audio, researchers have attempted to build large multi-modal models (LMMs) based on LLMs. Fuyu[16] directly fused image embedding vectors with text embedding vectors into the input model, and made preliminary attempts at semantic coherence between text and image data. LLaVa[17] connects an image encoder and a LLM through a projection matrix, trained to eliminate modeling illusions caused by text-image semantic mismatches. However, Cogvlm[18] indicated that direct alignment may not always be appropriate due to the fact that some details in the text do not exactly correspond one-to-one with those in the image, so Cogvlm processed the text and image inputs independently by means of different attention and full connectivity layers. In order to enable LLMs to cohesively handle multi-modal tasks, BLIP-2[19] proposes a Q-former module consisting of an image transformer and a text transformer, both of which achieve feature alignment through a shared self-attention layer and a cross-attention layer. The model was trained in two stages to obtain the image-text modal fusion, inference and generation capabilities. On this basis, MiniGPT-4[20] adopts the ViT-G module as image encoder from the pre-trained EVA-CLIP[21] and uses the Q-former as a multi-modal alignment module, which is connected to the LLM Vicuna[22] through a linear projection layer. These studies map the features of multi-modal data such as images into the feature space that is consistent with language, achieving the unified representation and fusion reasoning of multi-modal information. The above LMMs are pre-trained on a large number of general datasets and have relatively strong processing capabilities for general tasks, such as describing images of daily life. However, these LMMs lack professional knowledge in specific fields and it is difficult for them to directly solve highly professional problems like the description of structural damage images.

Currently, by learning professional knowledge in specific fields, LMMs have been widely applied in the research of professional fields such as medical diagnosis and industrial anomaly detection. CephGPT-4[23] built a high-quality dataset containing cephalometric analysis images and text data of doctor-patient dialogues. They fine-tuned models based on MiniGPT-4[20] and VisualGLM[24, 25] for cephalometric analysis and diagnostic dialogues. Myriad[26] is a LMM for industrial anomaly detection. It takes MiniGPT-4[20] as base model and adds visual expert modules, which can describe the defect features of industrial products in detail. These models are based on pre-trained open-source LMMs and fine-tune on the datasets in the vertical field, which provide a new method for civil engineering structural damage recognition. By inputting images and text and describing image details in the form of human-machine multi-round dialog.

General-purpose LMMs have undergone extensive pretraining on large scale image-text datasets, demonstrating strong comprehension abilities across both language and vision tasks. However, in the context of CE structures, damage occurrences during operation and maintenance are relatively infrequent, and the image collection process is cumbersome. As a result, images depicting structural damage appear less frequently in general datasets. Furthermore, the complex backgrounds associated with various damage types hinder the ability of general-purpose LMMs to acquire damage-related knowledge. While common damages such as surface cracks and holes are widely represented, more severe and less frequent issues, such as concrete spalling, exposed reinforcement, and corrosion, are underrepresented. This data imbalance limits the effectiveness of general-purpose LMMs in accurately identifying and describing structural damage.

To address these challenges, this study incorporates multi-modal chain-of-thought (CoT) reasoning to enhance the logical inference capabilities of structural damage identification models. Originally introduced in LLMs, CoT[27] enables models to decompose complex problems into sequential steps through user prompts, thereby improving reasoning performance. Zhang et al.[28] refined this approach by integrating language and vision modalities within a two-stage framework, effectively separating inference generation from answer reasoning to support a more structured reasoning process. However, CoT remains inherently unstable, as its effectiveness is fundamentally dependent on the model's reasoning capabilities. By constructing multi-turn dialogue datasets[29, 30], damage identification problems can be manually decomposed, enabling the model to learn high-quality reasoning processes and further enhance its inference capabilities. By establishing a multi-modal dataset for structural damage and fine-tuning general-purpose LMMs, it is possible to achieve a detailed description of the damage features through natural language.

This paper proposed SDIGLM, a structural damage identification LMM based on the open-source model VisualGLM-6B[24, 25], augmented with a multi-modal CoT framework. This paper construct a multi-modal dataset comprising images of apparent structural damage alongside textual descriptions. The model then leverages multi-modal CoT reasoning to perform damage identification and generate detailed textual descriptions. By fine-tuning the model with low-rank adaptation (LoRA), SDIGLM achieves an accuracy of 95.24% in structural damage recognition across various infrastructure types, including buildings, bridges, and roads. Furthermore, it can describe damage characteristics, such as hole size, crack orientation, and corrosion severity, through multi-turn dialogues, facilitating improved interpretation and decision-making in structural health monitoring.

## 2. Method

Existing structural damage identification models based on CV are often constrained in their ability to recognize only a limited number of damage types. These models typically classify damage categories or generate semantic segmentation maps that distinguish damage from background elements. However, they lack the capability to describe damage characteristics in a manner that facilitates human-computer interaction.

General-purpose LMMs were characterized by extensive parameters and pre-training, which exhibited strong generalization capabilities. Fine-tuning these models, rather than training from scratch, offers a more efficient pathway toward developing a structural damage identification system with enhanced adaptability. Additionally, LMMs bridge the modality gap between images and textual descriptions by performing vector embedding and alignment across multi-modal data, thereby enabling the generation of textual descriptions for structural damage images.

In this paper, taking the open-source LMM VisualGLM-6B[24, 25] as the framework, a brand-new large multi-modal model for structural damage identification, SDIGLM, is proposed. The model consists of modules such as U-Net[31], ViT-g, Q-former, and ChatGLM[32]. The U-Net module is responsible for receiving the input of structural damage images and outputting semantic segmentation maps that eliminate irrelevant backgrounds. The ViT-g module receives the original images of structural damage and the semantic segmentation maps, divides the images and maps them into vector encodings. The Q-former module then accomplishes the semantic alignment from image encoding to text encoding, enabling the LLM ChatGLM to process image encodings. After receiving user questions, ChatGLM conducts reasoning in combination with image encodings and generates responses. When using SDIGLM for structural damage identification, the model first generates a semantic segmentation map of the damaged area through the U-Net module to form a visual CoT. Then, it forms a Language CoT through the prompt words input by the user. Through the reasoning process of the multi-modal CoT, the model can accurately answer users' questions related to structural damage information in natural language.

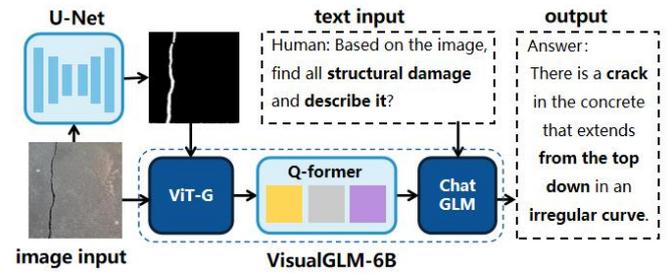

Fig. 1. Architecture of SDIGLM.

## 2.1 Vit-G and multi-modal data embedding

Since the image information of structural damage and the text information of natural language descriptions are two different data modalities, existing computer vision models are unable to understand the feature information of these two modalities simultaneously and thus cannot provide natural language descriptions for images. However, LLMs can segment natural language and encode it into high-dimensional space vectors. They also learn the correlations among these vector embeddings through a large amount of pre-trained text data, so that the text embeddings with similar semantics have close spatial distances. Therefore, LLMs possess the abilities to understand, reason about, and generate natural language. In order to enable models to process image data in the same way, it is necessary to first use an image encoder to uniformly represent the image information and text. Through a pre-trained image encoder, images can be converted into embeddings, making the image embeddings and text embeddings be in the same vector space. Then, through a pre-trained alignment module, the spatial distances of similar semantic embeddings in different modalities can be shortened, achieving the semantic integration of images and text. The image embedding and feature alignment methods are shown in Figure 2.

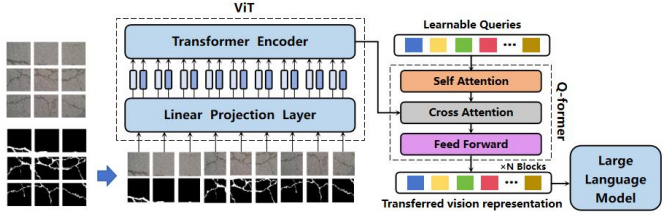

Fig. 2. Image data embedding and feature alignment methods.

The image encoder employs the ViT-G model. Firstly, it is necessary to divide the original structural damage images and semantic segmentation images into image patches, and then flatten each pixel of the image patches to obtain one-dimensional vectors. Suppose that both the resolution of the original structural damage images and the semantic segmentation images is $H \times W$, and the total number of channels is $C$. Divide them into image patches with a resolution of $P \times P$, and flatten each pixel of the image patches to obtain one-dimensional vectors with a length of $L = P^2 C$. Therefore, the structural damage images and the semantic segmentation images are converted into a vector sequence $T \in R^{N \times L}$.

The image patches are mapped into vector embeddings through a linear projection layer. And positional embeddings are added to each image patch to enhance its global perception ability and capture the overall content in the image. The calculation method is shown in Equation (1).

$$Z = [t_1 E, t_2 E ...... t_N E] + E_{pos} \quad (1)$$

where $E$ signifies the linear projection transformation, $E_{pos}$ represents the positional embedding of the image patch, which is learnable parameter. $t_i$ denotes the -$i$th vector of image embedding.

Image embedding feature extraction is carried out through the Transformer encoder[13] composed of multi-head self-attention(MSA) layers and feed-forward neural network(FFN) layers alternately. In the MSA layer, the input image patch embedding sequence undergoes linear transformations to obtain the Query, Key, and Value matrices for each head respectively, and the weights of these linear transformations are learnable matrices. Then, the dot product operation is performed on the Query matrix and the Key matrix of each head to calculate the similarity. Subsequently, scaling and normalization processing using the Softmax function are carried out to obtain the attention scores, which are then multiplied by the Value matrix to obtain the output of each head. Through such a multi-head self-attention mechanism, the model can capture the feature information in the input sequence and the dependency relationships among elements from multiple different representation subspaces. The outputs of each head are concatenated along the feature dimension and then input into the feed-forward neural network with the non-linear activation function ReLU for processing and transformation, and thus the image embedding after feature extraction is obtained. The calculation formulas are shown in Equations (2-5).

$$Q_i = Z W_Q^i, \quad K_i = Z W_K^i, \quad V_i = Z W_V^i \quad (2)$$

$$Z_i^{'} = A_i V_i = \text{Softmax}\left(\frac{Q_i K_i^T}{\sqrt{d_k}}\right) V_i \quad (3)$$

$$Z_{multi-head}^{'} = Z_{concat}^{'} W_z = \text{Concat}(Z_1^{'}, Z_2^{'} ...... Z_h^{'}) W_z \quad (4)$$

$$Z^{"} = [\text{ReLU}(Z^{'} W_1 + b_1)] W_2 + b_2 \quad (5)$$

where $Q_i$, $K_i$ and $V_i$ denote the Query, Key, and

Value matrices of the $i$-th head, respectively. $W_Q^i$, $W_K^i$ and $W_V^i$ correspond to the learnable linear transformation weight matrices of the $i$-th head, respectively. $d_k$ is the dimension of the Key matrix. $h$ is the number of heads. $A_i$ and $Z_i'$ refer to the attention score and the output of each head.

## 2.2 Q-former and multi-modal data alignment

In order to describe structural damage images using natural language, it is also necessary to align the image embeddings $Z''$ with the text embeddings to achieve semantic integration. The Q-Former module improves the model's ability to represent multi-modal information and the effect of information retrieval by introducing a learnable query embedding mechanism. Firstly, the query token embedding is mapped into query vectors for information retrieval through the self-attention layer. Then, the interaction between the query vectors and the input image embeddings is achieved in the cross-attention layer. The correlation between the query vectors and the input image embeddings is calculated through the dot product attention mechanism, and an embedding method that conforms to the image semantics is obtained through learning. This module is trained using a large amount of paired data of images and texts in the pre-training process to learn the embedding alignment relationship between images and texts.

The Q-Former module consists of a self-attention layer, a cross-attention layer, and a feed-forward neural network layer. The self-attention layer and the feed-forward neural network are similar to the methods mentioned above. In the cross-attention layer, the query vectors are learnable vectors, while the key matrix and the value matrix are obtained from the image feature embeddings. The cross-attention mechanism guides the model to retrieve information related to text semantics from the image features, which is used to obtain image feature embeddings with a spatial distance close to that of the text embeddings.

All of these modules can be pre-trained using general datasets so that the model can possess certain general capabilities, which helps to reduce the amount of training data for specific tasks such as structural damage identification. The pre-training methods can be referred to the relevant article[19]. Among it, the Q-Former module undergoes multi-objective pre-training on 129 million images in tasks such as image-text contrastive learning, image-grounded text generation, and image-text matching.

## 2.3 ChatGLM and natural language generation

In order to describe damaged images using natural language, the open-source LLM ChatGLM2-6B is used to receive text input, perform encoding inference, and generate language. ChatGLM2-6B needs to receive the user's text questions, segment the question text into tokens, and convert them into corresponding word vectors. These word vectors, along with the image encoding aligned with the Q-Former module, undergo feature extraction through the Transformer encoder. The autoregressive mechanism is utilized to generate tokens of the answer one by one. Meanwhile, the autoencoding mechanism ensures semantic consistency, ultimately producing an accurate and logical answer.

ChatGLM2-6B has completed pre-training on 1.4T identifiers and human preference alignment training on general-purpose datasets. It can perform natural language inference and generation in both Chinese and English. Based on the FlashAttention[33] technology, the context length of the model's dialogue can reach 32K. Meanwhile, due to the Multi-Query Attention[34] technology, ChatGLM2-6B features a more efficient inference speed and lower GPU memory consumption. Therefore, choosing ChatGLM2-6B as the basic language model can meet the requirements of the structural damage description task.

**2.4 U-net and multi-modal chain of thought reasoning**

Due to the complex environmental backgrounds where civil engineering structures are located, along with a wide variety of structural types and damage types, this has caused great difficulties for the damage identification and reasoning. It is hard for LLMs to extract the characteristic information of damages, which lead to model hallucinations. Effective reasoning thus is critic for the damage identification in such complex settings, where chain of thought (CoT) technique could help. CoT refers to the technique that guides the model to break down complex problems into solutions of step by step through the prompt, which can significantly improve the reasoning ability of model when dealing with complex problems. In this section, inspired by the text CoT prompt, a visual CoT is designed by invoking a U-net semantic segmentation model to enhance the accuracy of model damage identification and the logic of feature description in complex environments.

During the damage identification process, the model only needs to focus on the local damage features of the image, while the complex background information needs to be ignored. And the semantic segmentation maps that can be generated by existing computer vision models have excluded irrelevant background information and contain the locations and areas of damages. The model consists of two convolution layers and two transposed convolution layers. The structural damage images are normalized, and the normalized images are input into the down sampling encoder, which contains two convolution computations with 32 channels, a convolution kernel size of 3, and a stride of 1, and the ReLU activation function is used. The feature codes are then input into the up sampling decoder, which contains two transposed convolution computations with 32 channels, a transposed convolution kernel size of 3, and a stride of 1, and the ReLU activation function is also used. The calculation formula for each layer is as follows. The calculation method for each layer can be seen in Equation (6-7).

$$Y = \mathrm{Re}\,\mathrm{LU}(X_{image} * W_{conv}) \qquad (6)$$

$$VR = \mathrm{Re}\,\mathrm{LU}(Y \otimes W_{t-conv}) \qquad (7)$$

where $X_{image}$ is defined as the input structural damage image while $Y$ refers to the feature code output by the encoder. $*$ and $\otimes$ symbolize the convolutional and transposed convolutional computation respectively. $W_{conv}$ and $W_{t-conv}$ stand for the convolutional kernel and transposed convolutional kernel. $\mathrm{Re}\,\mathrm{LU}(x) = \max(0, x)$ is the activation function. $VR$ signifies the semantic segmentation map of the output damaged area and the background. The generation of semantic segmentation maps serves as a visual reasoning process to eliminate irrelevant information and avoid causing interference to damage diagnosis.

In the standard answer of the LMM, the model will generate answers based on the images and text questions input by users, generating a text embedding(token) each time. Equation (8) shows its reasoning process.

$$p(A| X_{text}, X_{image}) = \prod_{i=1}^{|A|} p_{LM}(a_i| X_{text}, X_{image}, a_{<i}) \quad (8)$$

where $X_{text}$ and $X_{image}$ are defined as the input question and image. $A$ stands for the answer generated by LMM. $a_i$ denotes the -$i$th token that makes up the answer.

In multi-modal CoT reasoning, the model will break down the questions according to the prompt words input by users and complete them step by step. In this section, the generated semantic segmentation map of damage is provided to the model as prior knowledge to form a visual thinking process, and the model is guided by text prompt words to first judge the damage category and then describe the damage features. Equations (9-11) represent the multi-modal CoT reasoning process of the LMM. It can be seen that by splitting the problem into intermediate processes through the multi-modal CoT, more prior knowledge can be provided for the final reasoning result. The multi-modal CoT reasoning process is shown in Figure 3.

$$p(A, R| X_{text}, X_{image}, I, VR) = p(A| I, X_{text}, X_{image}, VR, R) \cdot p(R| X_{text}, X_{image}, I, VR) \quad (9)$$

$$p(R| X_{text}, X_{image}, I, VR) = \prod_{i=1}^{|VR|} p_{LM}(r_i| X_{text}, X_{image}, I, VR, r_{<i}) \quad (10)$$

$$p(A| X_{text}, X_{image}, I, VR, R) = \prod_{j=1}^{|A|} p_{LM}(a_i| X_{text}, X_{image}, I, VR, R, a_{<j}) \quad (11)$$

where $I$ corresponds to the CoT prompt words. $VR$ symbolizes the semantic segmentation map, which is the content of visual reasoning. $R$ is the content of textual reasoning.

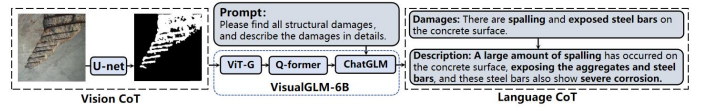

Fig. 3. The reasoning process of the multi-modal chain of thought.

## 3. Implementation

### 3.1 Multi-modal datasets

Although several general-purpose LMMs have demonstrated a certain degree of proficiency in image-text semantic alignment and text generation, their application in structural damage identification of CE structures in complex environment remains a challenge. Common types of damage, such as concrete cracks and road potholes, which are prevalent across various scenarios, can often be effectively recognized by these models. However, identification of less frequent damage types in CE structures, present a significant challenge. The complexity and diversity of background environments further exacerbate these limitations, potentially hindering the performance of general-purpose LMMs in specialized tasks.

Given these constraints, the development of domain-specific multimodal datasets tailored to civil engineering structural damage is paramount. Such datasets would enable the fine-tuning of LMMs to enhance their ability to detect and classify a broader spectrum of structural anomalies with greater accuracy. This targeted approach is crucial for advancing the applicability of LMMs in structural health monitoring tasks, where precision and reliability are critical.

The training datasets of LMMs consist of the matched images and text descriptions. This structure necessitates the collection of damage images from diverse structural scenarios, accompanied by detailed textual annotations that describe the features of the damage. This section selected images of concrete cracks and holes from CRACK500[35], GAPs384[36], and CFD[37] datasets, and images of pavement potholes from UDTIRI[38] dataset, and other damage images[39] such as concrete spalling, exposed steel bar, steel bar corrosion, cracks in the steel box girders of bridges and the peeling off of the protective layers of steel components. These damage images were processed through data cleaning and data enhancement such as rotation and folding. The text descriptions of damage categories, shapes, severity and other feature information are manually supplemented. The multi-modal dataset constitutes **11,722** sets of mutually matching original damage image and text description. Examples of the datasets are shown in Table 1. To ensure that the model can conduct correct reasoning and give proper answers to the given instructions, this section used predefined prompt words and response labels to standardize the training samples. The format of the training text consists of three parts {"img", "prompt", "label"}, where "img" refers to the corresponding image path. To enhance the model's reasoning ability, "prompts" are constructed in the form of multi-round historical conversations with question denoted as "Q" and answer denoted as "A". By splitting the questions into damage category critique and damage feature description, the model can obtain the inference ability of textual thinking through training.

Table 1. Examples of multi-modal structural damage datasets.

| structural damage images | text descriptions of damaged images |
|---|---|
| 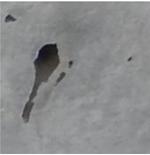 | {**"img"**: "traindata/00392.jpg", **"prompt"**: "Based on the picture, determine whether there is damage to the structure in the picture.", **"label"**: "There are holes in the concrete surface."} |
| 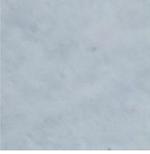 | {**"img"**:"traindata/00059.jpg",**"prompt"**:"Judge whether there is damage to the structure in the picture." ,**"label"**:"There is no obvious damage"} |
| 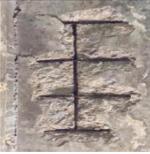 | {**"img"**:"traindata/01563.jpg",**"prompt"**:" **Q**: Determine whether there is damage to the structure in the picture. **A**: There are concrete spalling and exposed steel bars. **Q**: Please describe the characteristics of the damage in detail.",**"label"**:"A large amount of concrete on the surface of the structure has spalled off, causing the internal aggregates and steel bars to be exposed. Meanwhile, these steel bars have also suffered from serious corrosion problems."} |
| 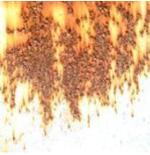 | {**"img"**:"traindata/03049.jpg",**"prompt"**:" **Q**: Please find all structural damage in the picture? **A**: The components in the picture are corroded. **Q**: Please describe the features of the damage based on the picture.",**"label"**:"The protective layer on the surface of the component has peeled off, causing the component to be exposed and suffer from serious corrosion damage."}, |
| 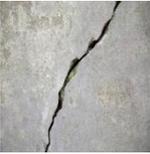 | {**"img"**:"traindata/05514.jpg",**"prompt"**:" **Q**: Determine whether there is damage to the structure in the picture. **A**:There are cracks on the surface of the concrete in the picture. **Q**: Please describe the characteristics of the damage according to the picture.",**"label"**:"An oblique crack expanding from the upper right to the lower left direction was observed on the surface of the concrete, and its path showed an irregular curve."}, |

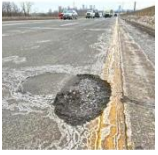
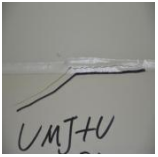

{**"img"**:"traindata/08076.jpg",**"prompt"**:"**Q**: Please find all structural damage in the picture. **A**: There is a potholes on the road in the picture. **Q**: Please describe the characteristics of the damage according to the picture.",**"label"**:"There is a pothole on the surface of the road, and the materials around it have cracked and been crushed."},

{**"img"**:"traindata/10031.jpg",**"prompt"**:"**Q**: Please check if there is any damage to the structure shown in the picture. **A**: There is a crack on the steel components. **Q**: Please describe the features of the damage",**"label"**:"On the surface of the steel component, a tiny crack is revealed in the area adjacent to the weld. This crack may seriously affect the stability of the structure."}

### 3.2 Fine-tuning

The base model of SDIGLM has been pre-trained on approximately 300 million image-text pairs. While this extensive pre-training grants the model strong capabilities in general image description tasks, its performance in domain-specific applications, such as structural health monitoring, remains constrained due to the absence of relevant damage-related data in its pre-training corpus. To bridge this gap, this section further fine-tune the model using the multi-modal structural damage datasets curated in Section 3.1.

For generating textual descriptions of structural damage images, the model takes both images and query texts as input, subsequently producing a descriptive textual output. To enhance the quality of these generated descriptions, the cross-entropy loss function is adopted as the optimization criterion. During text generation, the model sequentially produces word vector embeddings, referred to as tokens. In the training phase, the ground-truth tokens from the dataset serve as the correct labels, and cross-entropy loss quantifies the divergence between the predicted probability distribution and these labels. The formal computation of the cross-entropy loss is expressed in Equation (12).

$$Loss = -\sum_{i=1}^{N}\sum_{j=1}^{J-1} y_j^i \log P_w(a_j^i | X_{text}, X_{image}, I, VR, R, a_{<j}) \quad (12)$$

where $N$ denotes the number of samples and $J$ denotes the length of predicted sentence. $a_j^i$ represents the -$j$th token output by the -$i$th sample, while $y_j^i$ corresponds to the correct probability distribution of the -$j$th token in the -$i$th sample.

The proposed approach effectively quantifies discrepancies between the model-generated textual descriptions of structural damage and the ground-truth labels, facilitating parameter optimization and enhancing the model's performance in structural damage description tasks. However, SDIGLM consists of 7.8 billion parameters, making full-parameter training computationally expensive. Given the objective of enabling the model to learn damage identification from a small dataset while preserving core pre-trained knowledge, this section adopt a Low-Rank Adaptation (LoRA) strategy. By integrating LoRA into the model, only a subset of parameters is introduced and updated during training, significantly reducing computational overhead. The optimization objective for this adaptation is formally defined in Equation (13).

$$\Delta W = \arg\min_{\Delta W} Loss(W_0 + \Delta W) \quad (13)$$

where $\Delta W$ stands for the newly added learnable parameters (LoRA). The number of its parameters is much smaller than that of the original parameters $W_0$.

In order for general-purpose LMMs to maintain capabilities across diverse tasks, their parameter matrices must be high-dimensional. However, for domain-specific applications such as structural damage identification, only a fraction of the model's full capacity is required. To address this, LoRA introduces parallel learnable low-rank matrices into the neural network layers of the original model. As illustrated in Figure 4, these matrices are obtained by decomposing the original high-rank parameter matrix into the product of two smaller matrices. Following fine-tuning, the adapted parameters are combined with the original model parameters. The detailed computation process is provided in Equation (14).

$$h = W_0 x + \Delta W x = W_0 x + ABx \quad (14)$$
$$A \in R^{i \times r}, B \in R^{r \times o}$$

To reduce the rank of the original parameter matrix, it is factorized into matrices $A$ and $B$, both of which have narrower dimensions. The input dimension and output dimension are denoted as $i$ and $o$, And $r$ is defined as a hyperparameter denoting the rank of the parameter matrix, which is set to 10 in this study. Prior to training, $A$ is initialized to zero, while $B$ is randomly initialized using standard Gaussian distribution. This initialization ensures that the contribution of the LoRA adaptation remains neutral before training, preventing interference with the original model's pre-trained knowledge. This rank reduction strategy effectively minimizes the number of trainable parameters, making it well-suited for small-sample fine-tuning scenarios.

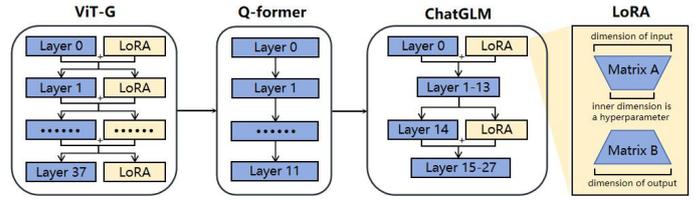

Fig. 4. Model fine-tuning methods based on low-rank adaptation.

Since structural damage images are underrepresented in the model's pre-training corpus, and the image encoder must embed both raw images and semantic segmentation maps, fine-tuning is applied to all parameters of the image encoder to enhance its ability to extract damage-related features. Additionally, to improve the accuracy and domain relevance of the generated descriptions, low-rank adapters are incorporated into the 0th and 14th layers of ChatGLM, while the remaining model components are kept frozen.

The fine-tuning of the SDIGLM model was implemented on an NVIDIA A10 GPU with 24GB of video memory, requiring 17 hours to complete 20 epochs on the constructed damage dataset.

## 4. Experiments and Results

To assess the effectiveness of SDIGLM in structural damage identification relative to general-purpose LMMs, this section utilize the SuperCLUE-V benchmark[40] as a reference. This benchmark evaluates LMMs through open-ended questions, assessing both fundamental and applied capabilities, thereby providing a comprehensive measure of a model's general multi-modal understanding. Among the top-performing models on this benchmark, GPT-4o and GLM-4v are selected as comparison baselines for SDIGLM.

Furthermore, to examine the contributions of multi-modal CoT reasoning and fine-tuning on structural damage datasets, this section also selects the base model VisualGLM-6B (base model), VisualGLM-6B(fine-tuned) that has been fine-tuned but not integrated with the multi-modal CoT, and SDIGLM proposed in this paper for ablation experiments. The details of these models are summarized in Table 2. The evaluation dataset consists of 210 images covering seven structural damage categories. All models are accessed via API and tested in a zero-shot setting.

Table 2. Experimental Model Setup.

|  | LMM | fine-tunned | multi-modal CoT |
|---|---|---|---|
| GPT-4o | √ |  |  |
| GLM-4V | √ |  |  |
| VisualGLM-6B (base model) | √ |  |  |
| VisualGLM-6B (fine-tunned) | √ | √ |  |
| SDIGLM(ours) | √ | √ | √ |

### 4.1. Qualitative Results

A qualitative assessment is conducted to evaluate the ability of SDIGLM to describe structural damage categories and characteristics in natural language. Figure 5 presents the damage identification results of SDIGLM. Given an input damage image, SDIGLM first generates a semantic segmentation map of the structural damage and progressively analyzes damage types and features through a multi-round dialogue guided by prompt words.

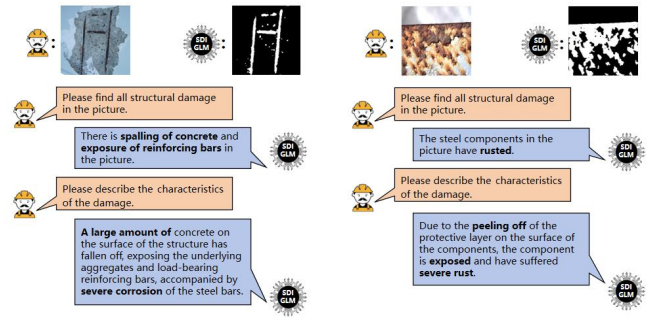

Fig. 5. Examples of structural damage identification by SDIGLM.

To verify the effectiveness of multi-modal CoT, Figures 6 and 7 compare SDIGLM with VisualGLM-6B, which lacks the U-Net module for generating multi-modal CoT. Figure 6 depicts a road pothole damage scenario, where SDIGLM first generates a semantic segmentation map, filtering out irrelevant background details such as vehicles. It then identifies and describes the damage category and characteristics through multi-turn interactions. Figure 7 presents an image of concrete holes, where VisualGLM-6B detects only a single deep hole, whereas SDIGLM, leveraging visual reasoning, accurately identifies multiple small holes as well as larger localized defects. These results demonstrate how multi-modal CoT enables synergy between LMMs and existing CV models, enhancing damage identification capabilities.

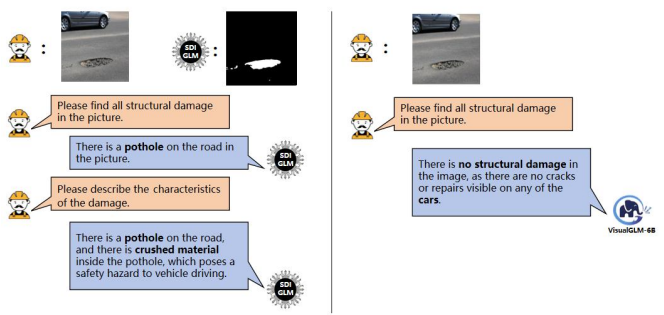

Fig. 6. Examples of the reasoning effect of the multi-modal chain of thought.

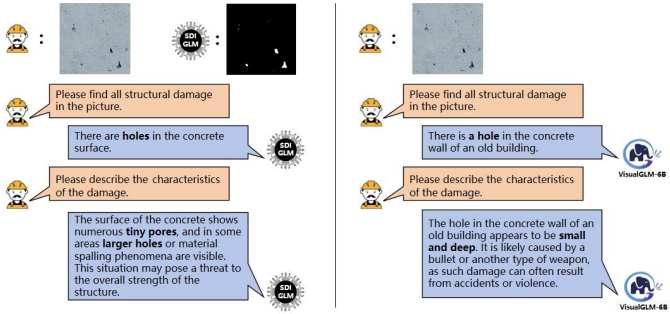

Fig. 7. Examples of the reasoning effect of the multi-modal chain of thought.

Further comparisons between SDIGLM and general-purpose LMMs in describing structural damage features are illustrated in Figure 8. The left-side example examines surface cracks in a steel box girder, where models trained on generic datasets often fail to distinguish structural defects from background elements. For instance, VisualGLM-6B(base model) misidentifies a weld seam as paint, and GPT-4o fails to detect fine cracks. In contrast, SDIGLM accurately identifies the cracks, describing both their location and potential impact. The right-side example evaluates a steel component with spalling and rusted protective coating, where VisualGLM-6B(base model) misinterprets rust as stains. SDIGLM, however, correctly identifies the spalling and corrosion, demonstrating its domain-specific adaptability.

Fig. 8. Comparison of damage identification among four LMMs.

## 4.2. Quantitative Results

A quantitative evaluation is conducted to assess model performance in damage category identification. To eliminate subjective bias, scoring is based solely on the correct identification of damage types, disregarding detailed descriptions and severity estimates. The dataset includes 210 structural damage images, covering seven damage categories. The number of correct classifications for each category and the overall accuracy are reported in Table 3.

Table 3. The identification accuracy of damage categories for each model.

| | SDIGLM (ours) | GPT-4o | GLM-4v | VisualGLM-6B (base model) | VisualGLM-6B (fine-tunned) |
|---|---|---|---|---|---|
| Road potholes | **30** | 28 | 29 | 23 | 27 |
| Concrete cracks | **30** | 30 | 29 | 29 | 30 |
| Concrete holes | **29** | 25 | 20 | 5 | 27 |
| Concrete spalling and rebar exposure | **28** | 25 | 14 | 9 | 25 |
| Steel elements cracks | **29** | 24 | 26 | 14 | 23 |
| Steel elements spalling and corrosion | **27** | 24 | 22 | 18 | 25 |
| undamaged | **27** | 27 | 18 | 15 | 20 |
| Accuracy | **95.24%** | 87.14% | 75.24% | 53.81% | 84.29% |

From Table 3, it is evident that VisualGLM-6B (base model) exhibits moderate accuracy in detecting common defects such as road potholes and concrete cracks. However, it struggles with more complex structural damages, including concrete spalling, rebar exposure, and steel element deterioration. Additionally, for intact structures, it occasionally hallucinates defects, leading to false positives. Similar limitations are observed in

GPT-4o and GLM-4v, though their higher parameter count improves accuracy relative to the base model.

Ablation experiments comparing VisualGLM-6B (base model), VisualGLM-6B (fine-tuned), and SDIGLM confirm that both fine-tuning and multi-modal CoT reasoning contribute significantly to performance enhancement. Fine-tuning allows the base model, despite its smaller parameter count, to match the accuracy of large-scale LMMs. Moreover, integrating multi-modal CoT reasoning enables SDIGLM to achieve the highest accuracy of 95.24%, surpassing general-purpose top-ranked LMMs in structural damage identification.

A comparative analysis of SDIGLM and existing computer vision-based structural damage identification models is presented in Table 4. Traditional CV models lack language-based reasoning capabilities, preventing them from generating textual descriptions of damage images. Additionally, most CV models are domain-specific, focusing solely on concrete or steel structures, thereby limiting their applicability across diverse structural damage types. In contrast, SDIGLM demonstrates multi-domain adaptability, effectively identifying various damage types across different structural materials.

The experimental results illustrate that through fine-tuning and multi-modal CoT reasoning, the 7.8-billion-parameter SDIGLM model surpasses general-purpose LMMs with over 100 billion parameters in structural damage identification.

Furthermore, SDIGLM can be locally deployed and executed on a 24GB GPU, significantly reducing the computational cost associated with running large-scale LMMs. This efficiency makes SDIGLM a practical solution for engineering applications, facilitating on-site damage assessment and real-time decision-making.

Table 4. Comparison of damage identification capabilities.

| | SDIGLM | VGGNet[41] | ZFNet[42] | GoogLeNet[43] |
|---|---|---|---|---|
| model architecture | LMM | CNN | CNN | CNN |
| language ability | √ | - | - | - |
| categories of damage identification: | | | | |
| road potholes | √ | - | - | - |
| concrete cracks | √ | √ | - | √ |
| concrete holes | √ | - | - | - |
| concrete spalling and rebar exposure | √ | - | - | √ |
| steel elements cracks | √ | - | - | - |
| steel elements spalling and corrosion | √ | - | √ | - |
| undamaged | √ | √ | √ | √ |
| Accuracy | 95.24% | 97.07% | 96.11% | 94.44% |

## 5. Conclusion

This study addresses two critical limitations in current computer vision (CV)-based structural damage recognition systems: 1) their restricted capacity to identify a comprehensive range of damage types, and 2) their inability to generate natural language descriptions of damage characteristics. To overcome these limitations, this study propose SDIGLM, an advanced LMM framework built upon the VisualGLM-6B architecture. The proposed framework incorporates a U-Net-based semantic segmentation module to generate damage segmentation maps, which serve as visual CoT. A specialized multi-round dialogue fine-tuning dataset was developed to enhance the model's logical reasoning capabilities, complemented by a language CoT implemented through domain-specific prompt engineering. Through the synergistic integration of multi-modal CoT, SDIGLM demonstrates superior performance compared to general-purpose LMMs in structural damage identification tasks, achieving a accuracy of 95.24% across diverse structure types. Additionally, SDIGLM provides detailed damage characterizations, such as hole dimensions, crack orientations, and rust severity, demonstrating the potential of LMMs in civil engineering applications. By aligning image embeddings with textual semantics, SDIGLM lays the groundwork for automated inspection report generation and decision support in structural maintenance. However, it is important to note that real-world civil infrastructure presents highly diverse and complex conditions, which are difficult to comprehensively capture in a single dataset. Continuous data collection from real-world applications and periodic model updates are essential for further enhancing SDIGLM's capabilities and ensuring its long-term reliability in structural health monitoring.